\documentclass{article}
\usepackage{ijcai24}

\usepackage{times}
\usepackage{soul}
\usepackage{url}
\usepackage[hidelinks]{hyperref}
\usepackage[utf8]{inputenc}
\usepackage[small]{caption}
\usepackage{graphicx}
\usepackage{amsmath}
\usepackage{amsthm}
\usepackage{booktabs}
\usepackage{algorithm}
\usepackage{algorithmic}
\usepackage[switch]{lineno}
\usepackage{natbib}
\usepackage{multirow}
\usepackage{enumitem}
\setlist{nolistsep}
\usepackage{xcolor}
\usepackage{diagbox}
\usepackage{makecell}
\usepackage{multirow}
\usepackage{cite}
\usepackage{booktabs}
\usepackage{caption}

\usepackage{amsmath,amssymb,amsfonts}
\usepackage{xcolor}
\usepackage{listings}
\usepackage{algorithm}
\usepackage{algorithmic}
\usepackage{bbding}
\usepackage{newfloat}
\usepackage{listings}
\DeclareCaptionStyle{ruled}{labelfont=normalfont,labelsep=colon,strut=off} 
\floatstyle{ruled}
\newfloat{listing}{tb}{lst}{}
\floatname{listing}{Listing}

\title{Modality-Aware and Shift Mixer for Multi-modal Brain Tumor Segmentation}

\author{
Zhongzhen Huang$^{1,2}$
\and
Linda Wei$^1$\and
Shaoting Zhang$^{2}$ \Envelope \And
Xiaofan Zhang$^{1,2}$
\affiliations
$^1$Shanghai Jiao Tong University 
$^2$Shanghai AI Laboratory\\
\emails
\{huangzhongzhen, lullcant, xiaofan.zhang\}@sjtu.edu.cn,
zhangshaoting@pjlab.org.cn
}

\begin{document}

\maketitle

\begin{abstract}
Combining images from multi-modalities is beneficial to explore various information in computer vision, especially in the medical domain. As an essential part of clinical diagnosis, multi-modal brain tumor segmentation aims to delineate the malignant entity involving multiple modalities. Although existing methods have shown remarkable performance in the task, the information exchange for cross-scale and high-level representations fusion in spatial and modality are limited in these methods. In this paper, we present a novel \textbf{M}odality \textbf{A}ware and \textbf{S}hift \textbf{M}ixer (\textbf{MASM}) that integrates intra-modality and inter-modality dependencies of multi-modal images for effective and robust brain tumor segmentation. Specifically, we introduce a Modality-Aware module according to neuroimaging studies for modeling the specific modality pair relationships at low levels, and a Modality-Shift module with specific mosaic patterns is developed to explore the complex relationships across modalities at high levels via the self-attention. Experimentally, we outperform previous state-of-the-art approaches on the public Brain Tumor Segmentation (BraTS 2021 segmentation) dataset. Further qualitative experiments demonstrate the efficacy and robustness of MASM.

\end{abstract}

\section{Introduction}
\begin{figure}[t]
\centering
\scalebox{1}{
\includegraphics[width=\linewidth]{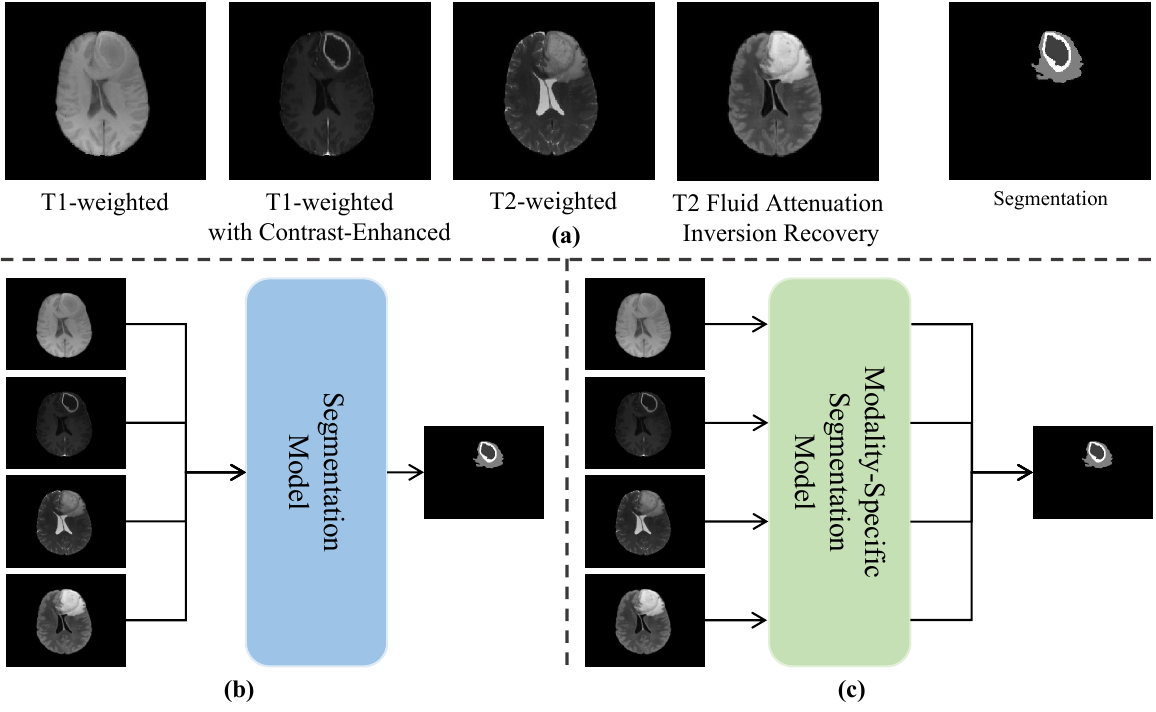}
}
\vspace{-10pt}
\caption{\textbf{(a)} Multi-modal brain tumor segmentation; \textbf{(b)} Early-Fusion strategy; \textbf{(c)} Later-Fusion strategy}
\vspace{-10pt}
\label{fig:intro}
\end{figure}

Leveraging images from multi-modalities has shown promising potential in real-world scenarios due to the contribution of various information, especially in the medical domain, where multi-modal medical images are utilized to delineate anatomical structures and other abnormal entities. For instance, the Computed Tomography (CT) plain scan can be used to evaluate morphology and detect abnormalities, and the contrast-enhanced CT scan assesses the blood supply to potential tumors, aiding in distinguishing between benign and malignant lesions. Moreover, there are several Magnetic Resonance Imaging (MRI) sequences, such as T1-weighted (T1), T1-weighted with contrast-enhanced (T1-CE), T2-weighted (T2), and T2 Fluid Attenuation Inversion Recovery (T2-FLAIR) are combined to emphasize and distinguish different tissue properties and areas of tumors, as shown in Figure~\ref{fig:intro}(a). As the most common cancer worldwide, elaborating on the characterization of brain tumors is vital for studying tumor progression and pre-surgical planning. However, the diagnosing process is time-consuming and error-prone. Thus, developing robust models for multi-modal brain tumor segmentation is of primary importance for current clinical diagnosis.

In recent years, promising advances in brain tumor segmentation have been primarily driven by the powerful capabilities of deep learning-based techniques. Convolution Neural Networks (CNNs) with the encoder-decoder architecture~\citep{kong2018hybrid,chen2019s3d,isensee2021nnu,myronenko20193d,jiang2020two}, which demonstrates state-of-the-art performances on various benchmarks, is an effective and reproducible solution for brain tumor delineation. Due to the capability of learning long-range dependencies, some transformer-based models~\citep{chen2021transunet,hatamizadeh2022unetr,hatamizadeh2022swin} have been exploited for powerful relation modeling. Despite the remarkable performances, how to explore and utilize the relationships among different modalities was not specifically designed, making it bleak for further promoting the accuracy of multi-modal images. 
Previous methods adopted an early-fusion strategy where multi-modal images are merely concatenated at the input and processed jointly as a single stream of the network, as illustrated in Figure~\ref{fig:intro}(b). However, such early-fusion methods overlooked the relationships among multiple modality inputs which is essential for clinical diagnosis. 

Instead of concatenating all the modalities in the early stage, our community has witnessed wide attempts to explore fusing features from multiple modalities in the later stage recently, as shown in Figure~\ref{fig:intro}(c). However, as mentioned in \citep{srivastava2012multimodal}, it can hardly discover highly non-linear relationships between different modalities, especially when the original T1, T2, T1-CE, and FLAIR images have different statistical properties due to significantly different image acquisition processes. To alleviate this problem, another volume of work~\citep{lin2022ckd, zhang2021modality, huang2022evidence, xing2022nestedformer} began to design specific fusion modules for multi-model information exchange and feature combination.  Nevertheless, such designs are still sub-optimal for multi-modal image segmentation for the following reasons. First, information exchange for multi-scale, which is vital for segmentation tasks, is not fully considered. Second, how to effectively fuse higher-level representations in spatial and modality, which is crucial for multi-modal images, is still not taken into account.


Following these premises, we leverage the power of transformers and introduce a novel method dubbed as \textbf{M}odality \textbf{A}ware and \textbf{S}hift \textbf{M}ixer (\textbf{MASM}) as shown in Figure~\ref{fig:Network}, which incorporates multi-scale context modeling and multi-modal relationships mining for accurate segmentation. To model multi-modal features at low levels, we propose a Modality-Aware module for more effective and reasonable information exchange across different modalities. Since different MRI sequences are often combined for diagnosis (e.g., T2 and FLAIR are together to observe water signals), the Modality-Aware module is carefully designed according to neuroimaging studies and attempts to establish the inter-modality dependencies between the specific pair of modalities. For multi-modal features at high levels, a transformer-based Modality Shift module with specific mosaic patterns is introduced to explore the complex relationships between different modalities better. We endow transformers with the capability of modalities modeling without additional parameters and computational costs by shifting patches along the modality dimension for later self-attention. For each modality, its patches are replaced with patches from other modalities according to the patterns. We further utilize the multi-modal features from all layers for the segmentation results.

%


Our contributions can be summarized as follows:
\begin{itemize}[noitemsep, topsep=0pt]
\item We propose a novel model that exploits the relationship and interaction among multiple modality inputs for multi-modal brain tumor segmentation.
\item The introduced Modality-Aware module enables reasonable information exchange on the multi-scale features according to neuroimaging studies, which are crucial in medical multi-modal image segmentation.
\item The Modality-Shift module is developed to explore the high-level features of multi-modalities. With the specific design,  this module operates without additional parameters or computational overhead.
\item We validate the effectiveness of our method in the BraTS 2021 Challenge. MASM achieves state-of-the-art performance on the benchmark compared to the universal and specifically designed multi-modal methodologies.
\end{itemize}
\section{Related Work}
\subsubsection{Medical Image Segmentation:} Convolutional Neural Networks (CNNs) have demonstrated significant effectiveness in medical image segmentation tasks \citep{ronneberger2015u}.  However, due to the local property of the convolutional kernels, the CNN-based segmentation models cannot learn long-range dependencies, which can severely impact the accurate segmentation of tumors that appear in various shapes and sizes. To cope with such an issue, another volume of transformer-based models has been exploited for powerful relation modeling. Chen \textit{et al.}~\citep{chen2021transunet} proposed TransUNet, which introduced the self-attention mechanism to model the global context for high-level features. After Vision Transformer (ViT) \citep{ViT} was shown to be a good visual feature extractor, there was a volume of new segmentation frameworks based on ViT. As a roadmap of utilizing a ViT as its encoder without relying on a CNN-based feature extractor, UNETR \citep{hatamizadeh2022unetr} has shown good performance in segmentation. Since multi-scale features play a pivotal role in medical image segmentation tasks such as tumor segmentation, a model is required to handle features across multiple scales effectively. To leverage the multi-scale features, SwinUNERT \citep{hatamizadeh2022swin} was proposed to compute self-attention in an efficient schema. In this paper, we try reasonable and efficient modules to boost the performance of models in the brain tumor segmentation task.

\subsubsection{Multi-modal Segmentation:} In medical fields, it is often necessary to use MRI scans to identify and locate brain tumors. However, a single MRI sequence, such as T1-weighted or T2-weighted images, may not provide sufficient information for accurate and robust segmentation results. Therefore, multi-modal segmentation methods should be employed, where the T1, T2, FLAIR, and T1-CE sequences are utilized concurrently according to the real clinical scenario. These sequences offer complementary information about the tumor's location, size, and other characteristics, thereby enhancing the accuracy of the segmentation. Recently, our community has witnessed a wide adoption of deep-learning techniques to model the relationships among multi-modal images for medical image segmentation tasks. Lin \textit{et al.} \citep{lin2022ckd} proposed a clinical knowledge-driven model with a dual-branch hybrid encoder that splits the modalities into two groups based on the imaging principle as input. MAML \citep{zhang2021modality} and MMEF \citep{huang2022evidence} proposed modality-specific models to extract different high-level representations and adopted a fusion module for multi-model information exchange and feature combination. Xing \textit{et al.} \citep{xing2022nestedformer} performed nested multi-modal fusion for different modalities by establishing intra- and inter-modality coherence to build the long-range spatial dependencies across modalities. Instead of applying one modality-fusion module for multi-scale features, we propose two novel modules (i.e. Modality-Aware and Modality-Shift) for exploiting the relationships among multiple modalities across different scales.
%

\section{Method}
\label{method}
\begin{figure*}[h]
\centering
\scalebox{0.9}{
\includegraphics[width=\linewidth]{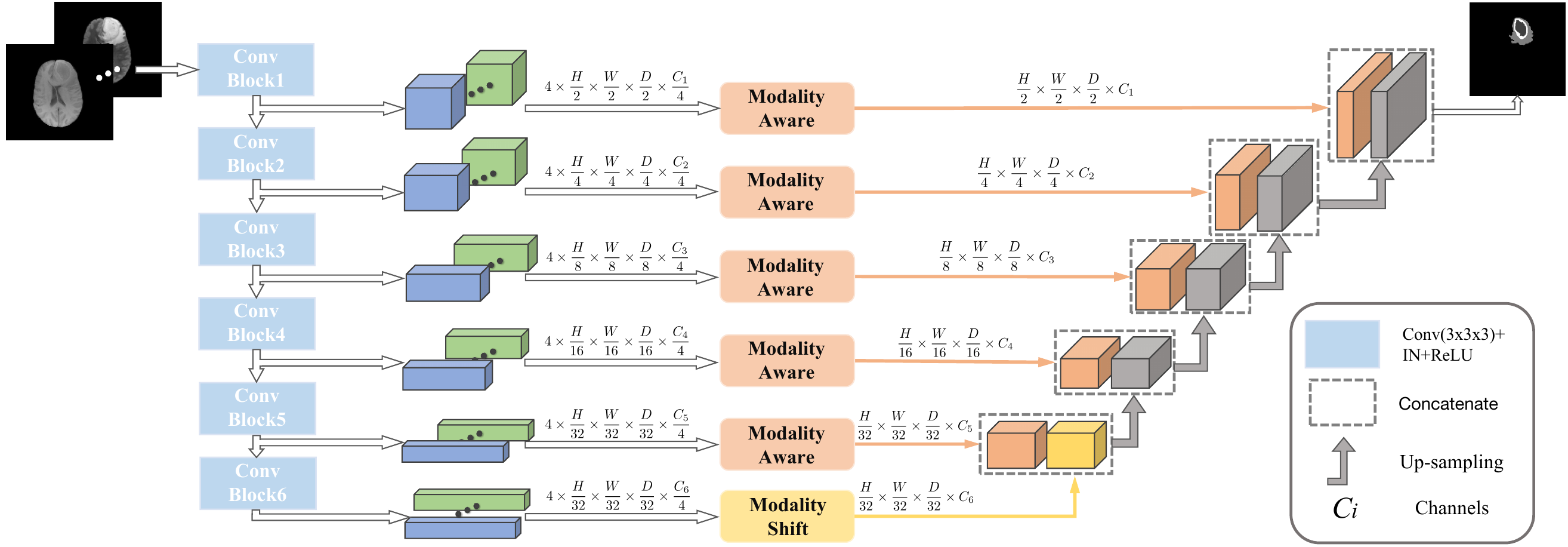}
}
\caption{The overall architecture of \textit{MASM} based on U-Net: 1) Modality-Aware module that handles the interaction between specific modalities, 2) Modality-Shift module that exploits high-level relationships among modalities.}
\vspace{-10pt}
\label{fig:Network}
\end{figure*}
As illustrated in Figure~\ref{fig:Network}, our MASM consists of a backbone based on U-Net, the Modality-Aware module, and the Modality-Shift module. We apply the Modality-Aware module and the Modality-Shift module in the original skip connection for modeling the multi-modal relationships. Features from the first five layers are regarded as low-level features where the Modality-Aware is adopted, while the Modality-Shift module is only utilized in the last layer for modeling high-level features. Our experiments reveal the effectiveness of such a design. Given multi-modal images $M^i \in \mathbb{R}^{D\times H \times W}, i \in [1,4]$, our model aims to output the segmentation results $S \in \mathbb{R}^{D\times H \times W \times 3}$, where $M^i$ represents modal T2, T1, T1-CE, and FLAIR respectively.
\subsection{Backbone}
Since U-Net~\citep{ronneberger2015u} has been proven powerful in medical image segmentation, we adopt it as the backbone to extract multi-scale features from multi-modal magnetic resonance images. Instead of employing the modality-specific encoder to extract features for each modality, we leverage a single shared encoder to represent the image features. The input of each modal image individually goes through the shared module, and this design can effectively reduce the number of model parameters. It is worth noting that the parameter sharing in the encoder does not degrade the performance. In a similar way, we can formulate the extracted feature maps as: 
$$
F^i_{j} \in \mathbb{R}^{d_j \times h_j \times w_j \times \frac{C_j}{4}}$$
where $d_j,h_j ,w_j={\frac{D}{2^j},\frac{H}{2^j},\frac{W}{2^j}}, j\in[1,5]$. And the high-level embeddings from the last layer are obtained as $ F^i_{6} \in \mathbb{R}^{ \frac{D}{2^5}  \times\frac{H}{2^5} \times \frac{W}{2^5} \times\frac{C_6}{4}}$.

\begin{figure*}[htp]
\centering
\scalebox{1}{
\includegraphics[width=\linewidth]{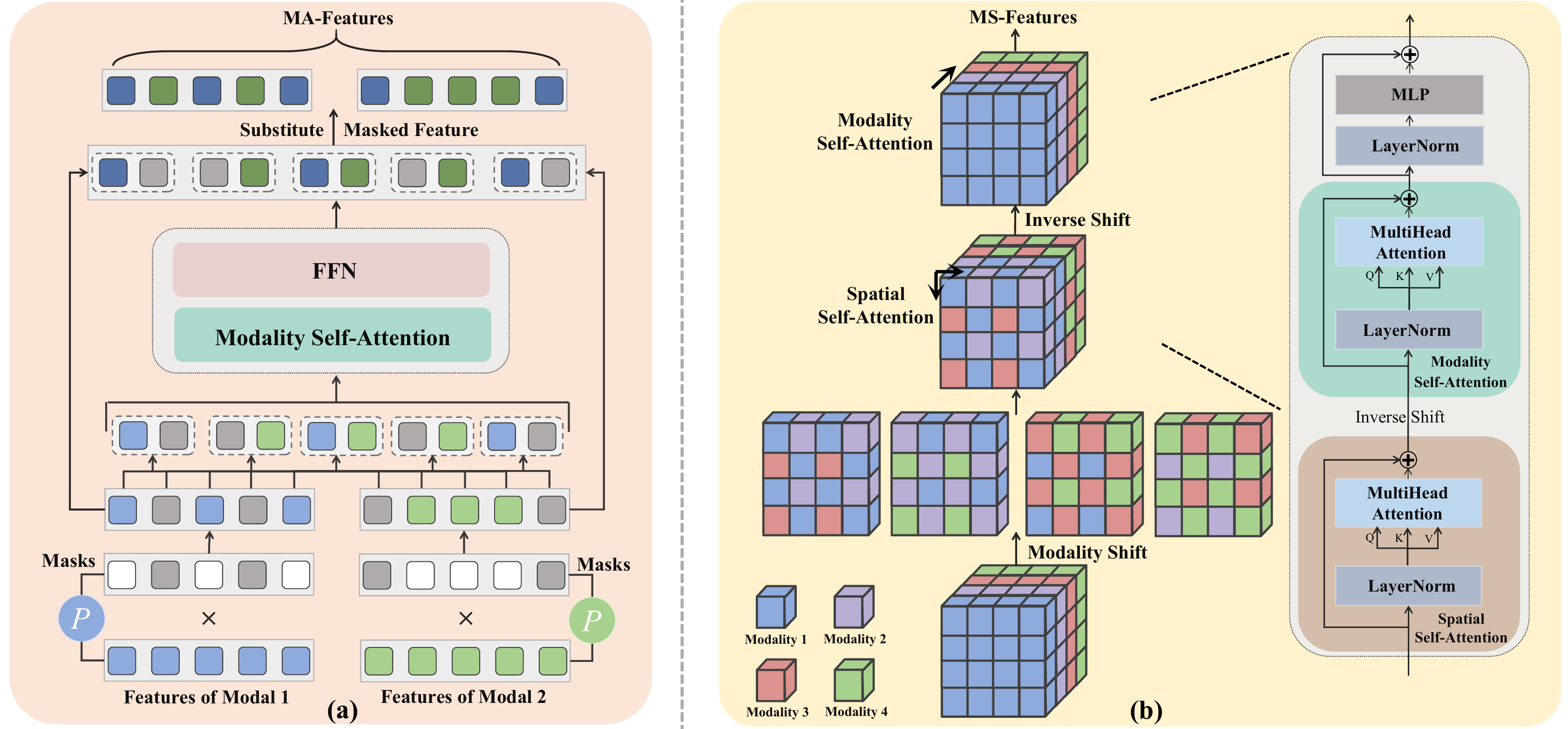}
}
\caption{\textbf{(a)} The illustration of Modality-Aware; \textbf{(b)} The process of Modality-Shift. The Modality-Aware is performed on two modalities (i.e., T2 and FLAIR), and The Modality-Shift is applied to three modalities (i.e., T1, T1-CE, and FLAIR).}
\vspace{-10pt}
\label{fig:aware_shift}
\end{figure*}

\subsection{Modality Aware Module}
In clinical diagnosis, T1 images are usually combined with T1-CE images to get valuable information about the presence and nature of various abnormalities, and T2 images are often combined with FLAIR images for detecting different types of information about the tissues. Given features $F^i_{L}$ in the layer $L$, we propose a Modality-Aware module to aggregate the features in a reasonable way, which models the relationships according to the neuroimaging studies to make the segmentation process conformance to the real scenario. 

Moreover, the presence of redundant patches in each modality feature is a common occurrence for volumetric medical images, especially for multi-modal segmentation, where normal patches repeatedly appear in multiple modalities of images. Redundant patches in each modality feature may undermine the process of information exchange. Therefore, we selectively mask the uninformative tokens to obtain more discriminative features and reduce the computational scope. The Modality-Aware module produces a binary decision mask for each scale feature to decide which patches are redundant and can be pruned for each modality and substituted by alignment features from other modalities after feature mixing. As shown in the left part of Figure~\ref{fig:aware_shift}, we first flatten each feature $F^i_{L}$ into a sequence  $\hat{F}^i_{L} \in \mathbb{R}^{N_L \times \frac{C_L}{4}}$, and binary decision masks $\mathbf{D}_i \in\{0,1\}^{N_L}$  is maintained to indicate whether to mask each token or not, where $N_L=d_L \times h_L \times w_L$ is the number of the sequence embeddings. We input $\hat{F^i_{L}}$ to the mask prediction module and compute the decision mask $\mathbf{D}$ as follows:
\begin{gather}
    \mathbf{y}_i^{\text {local }}=\operatorname{MLP}(\hat{F}^i_{L}) \in \mathbb{R}^{N_L \times C^{\prime}}\\
\mathbf{y}_i^{\text {global }}=\operatorname{Agg}(\operatorname{MLP}(\hat{F}^i_{L})) \in \mathbb{R}^{C^{\prime}},
\end{gather}
where $C^{\prime}$ is half the dimension of $\hat{F}^i_{L}$.  $\mathbf{y}_i^{\text {local}}$ and $\mathbf{y}_i^{\text {global}}$ are the local and global features computed by MLP~\citep{dosovitskiy2020image}  $\operatorname{Agg}$ is for aggregating the information from local features and can be implemented by an average pooling.

Intuitively, the local feature represents the information of each patch, and the global feature contains the context information. The global features will be expanded to the same length as local features and then concatenated with local features in the last dimension. Then, the concatenated vector is used to decide whether to mask the token:
\begin{gather}
\mathbf{y}_{ik}=\left[\mathbf{y}_{ik}^{\text {local }}, \mathbf{y}_{ik}^{\text {global }}\right], \quad 1 \leq k \leq N_L, \\
\boldsymbol{\pi}_i=\operatorname{Softmax}(\operatorname{MLP}(\mathbf{y}_i)) \in \mathbb{R}^{N_L \times 2},
\end{gather}
Since it is non-differentiable to get a mask $\mathbf{D}$ sampling from $\boldsymbol{\pi}$, following~\citep{rao2021dynamicvit}, we adopt the Gumbel-Softmax technique~\citep{jang2016categorical} to sample from $\boldsymbol{\pi}$ :
\begin{equation}
\mathbf{D}_i=\operatorname{Gumbel-Softmax}(\boldsymbol{\pi}_i)_{*, 0} \in\{0,1\}^{N_L},
\end{equation}
where index 0 in $\mathbf{D}$ represents masking the corresponding patch, Gumbel-Softmax is differentiable for end-to-end training.  After attaining the masks, $\hat{F}^i_{L}$ can be pruned by the Hadamard product with $\mathbf{D}_i$:

\begin{equation}
\tilde{F}^i_{L}\leftarrow \hat{F}^i_{L} \odot \mathbf{D}_i,
\end{equation}

As shown in Figure~\ref{fig:aware_shift}, Modality Self-Attention is utilized to compute the correlation and interaction between the modalities $\tilde{F}^i_{L}$. 
We use $\tilde{F}^1_{L}$ and $\tilde{F}^4_{L}$ (i.e., T2 and FLAIR) as an example to illustrate the Modality Self-Attention. In this module, each image patch of $\tilde{F}_{L}^i$ (e.g.  $\tilde{F}_{L1}^i$) is considered as one token, and tokens at the same location in different modalities form a paired modality sequence like $\tilde{\mathbf{z}}_1=\{\tilde{F}^1_{L1},\tilde{F}^4_{L1}\}$. Then the Modality Self-Attention is employed to the formed patches sequence $\tilde{\mathbf{z}}=[\tilde{F}^1_{L},\tilde{F}^4_{L}]$. This operation can be performed efficiently by reshaping the input $\tilde{\mathbf{z}}$ with batch size $B$ from $\mathbb{R}^{B \times 2 \times N_L  \times \frac{C_L}{4}}$ to $\mathbb{R}^{B \cdot N_L \times 2  \times \frac{C_L}{4}}$ and leveraging the self-attention~\citep{vaswani2017attention} scheme to compute the correlation for the above sequences:
\begin{gather}
Q=\tilde{\mathbf{z}} \mathrm{~W}^Q, K=\tilde{\mathbf{z}}\mathrm{~W}^K, V=\tilde{\mathbf{z}} \mathrm{~W}^V \\
\operatorname{Modality-SA}\left(\tilde{\mathbf{z}}\right)=\operatorname{Softmax}\left(\frac{Q K^T}{\sqrt{d}}\right) V \\
H_L = \operatorname{FFN}(\operatorname{Modality-SA}\left(\tilde{\mathbf{z}}\right)) \\
\operatorname{FFN}(x)=\max \left(0, x \mathrm{~W}_{0}+\mathrm{b}_{0}\right) \mathrm{W}_{1}+\mathrm{b}_{1}
\end{gather}
where $ \mathrm{~W}^Q, \mathrm{~W}^K, \mathrm{~W}^V \in \mathbb{R}^{d \times d}$, $\mathrm{W}_{0}\in\mathbb{R}^{d \times 4 d}$ and $\mathrm{W}_{\mathrm{1}} \in \mathbb{R}^{4 d \times d}$ are learnable parameters and  $b_{0}, b_{1}$ are the bias terms. In our implementation, $d$ is equal to $\frac{C_L}{4}$.

After getting the output $H_L$, we reshape it back to $\mathbb{R}^{B \cdot N_L \times 2  \times \frac{C_L}{4}}$ and attain $[h^1_L, h^4_L]$. Then, these masked features in $h^i_L$ are substituted with the corresponding token in $h^j_L$ as shown in Figure~\ref{fig:aware_shift} (a). For example, $\tilde{F}_{L3}^1$ is masked, we replace $h^1_{L3}$ with $h^1_{L4}$. The obtained features are concatenated together to generate ${F}'_{L}$ for the decoding stage.


\subsection{Modality Shift Module}
According to ~\citep{nie2016fully}, later fusion of high-level features is better for the complex relationships between different modalities. Inspired by the notable performance of transformers in modeling relationships between different entities, we endow transformers with the capability of modality modeling without additional parameters and computational costs. To explore complex relationships among modalities in high-level, we introduce a Modality-Shift Module to fuse the highest level features $F^i_{6}$ as shown in the right part of Figure~\ref{fig:aware_shift}. 

Specific mosaic patterns are designed for patch shifting along the modality dimension. We can define a generic modality shift operation in transformers as follows:
\begin{gather}
\mathbf{Z}^{1} =\left[\mathbf{z}^{1}_0,\mathbf{z}^{1}_1, \ldots, \mathbf{z}^{1}_N\right]\\
\mathbf{Z}^{2} =\left[\mathbf{z}^{2}_0, \mathbf{z}^{2}_1, \ldots, \mathbf{z}^{2}_N\right]\\ 
\mathbf{A} =\left[\mathbf{a}_0, \mathbf{a}_1, \ldots, \mathbf{a}_N\right] \\
\mathbf{Z}  =\mathbb{I}_{\mathbf{A}=\mathbf{1}}  \odot \mathbf{Z}^{1}+\mathbb{I}_{\mathbf{A}=\mathbf{2}} \odot \mathbf{Z}^2
\end{gather}
where $\mathbf{Z}^1,\mathbf{Z}^{2}$ represent the patch features for modality $1$ and modality $2$, respectively. $N$ is the number of patches, and $\mathbf{A}$ represents the matrix of shifting with $\mathbf{a}_i \in \{1,2,3,4\}$ indicating the source of the shifting patch $i$. $\mathbb{I}$ is an indicator asserting the subscript condition. ${\mathbf{Z}}$ is the output image patches after shift operation.

Using the proposed modality shift operation,  we can achieve information exchange among modalities at a high level. Since one specific modality has interacted with the other in the Modality-Aware module, we apply shift operation in the rest of the modality features in our case. To reduce the mixing space, we adopt fixed shift patterns. Namely, there is an invariant $A_i$ for each $M_i$. For example, $M^1$ is interacted with $M^4$, thus, $M^1$ would be fused with $M^2$ and $M^3$ by the shift operation as follows:
\begin{equation}
\tilde{F}^1_{6} =\mathbb{I}_{\mathbf{A}_1=\mathbf{1}}  \odot \hat{F}^1_{6}+\mathbb{I}_{\mathbf{A}_1=\mathbf{2}} \odot \hat{F}^2_{6}+\mathbb{I}_{\mathbf{A}_1=\mathbf{3}} \odot \hat{F}^3_{6}
\end{equation}
where $\hat{F}^i_6 \in \mathbb{R}^{N_6\times C_6}$ is the flatten feature. 

Then, we can exploit the complex relationships among modalities via the attention mechanism. The spatial self-attention and modality self-attention are employed sequentially in this module. The spatial self-attention can be performed by reshaping the input $\tilde{F}^i_6$ with batch size $B$ from $\mathbb{R}^{B \times 4 \times N_6  \times \frac{C_6}{4}}$ to $\mathbb{R}^{B \cdot 4 \times N_6   \times \frac{C_6}{4}}$. Different from the Modalitiy-Aware module, we adopt MultiHead Attention (MHA) and LayerNorm (LN) in this module, the process can be formulated as:
\begin{gather}
\label{eq:MHA}
\hat{X}= \operatorname{MHA}\left(\operatorname{LN}(X)\right)+ X \\ 
\operatorname{MHA}(x)=\left[\operatorname{Att}_{1}(x) , \ldots , \operatorname{Att}_{n}(x)\right] \mathrm{W}^{\mathrm{O}}
\\\operatorname{Att}_{i}(x)=\operatorname{Softmax}\left(\frac{x \mathrm{~W}_{i}^{\mathrm{Q}}\left(x \mathrm{~W}_{i}^{\mathrm{K}}\right)^{T}}{\sqrt{d_{n}}}\right) x \mathrm{~W}_{i}^{\mathrm{V}}
\end{gather}
where $X$ denotes the input features. $\mathrm{W}_{i}^{\mathrm{Q}}, \mathrm{W}_{i}^{\mathrm{K}}, \mathrm{W}_{i}^{\mathrm{V}} \in$ $\mathbb{R}^{d \times d_{n}}$ and $\mathrm{W}^{\mathrm{O}} \in \mathbb{R}^{d \times d}$ are learnable parameters, $d_{n}=d / n$, $[\cdot , \cdot]$ stands for concatenation operation.
After self-attention, patches from different modalities are shifted back to their original locations as follows:
\begin{equation}
\tilde{S}^1_{6} =\mathbb{I}_{\mathbf{A}_1=\mathbf{1}}  \odot {S}^1_{6}+\mathbb{I}_{\mathbf{A}_2=\mathbf{1}} \odot {S}^2_{6}+\mathbb{I}_{\mathbf{A}_3=\mathbf{1}} \odot {S}^3_{6}
\end{equation}
where $S^i_6$ is the output of the spatial self-attention and $\tilde{S}_{6} \in \mathbb{R}^{B \times 4 \times N_6  \times \frac{C_6}{4}}$ is the visual feature after shifting back.
Then, the modality self-attention is utilized to augment the feature fusion among modalities further. Similar to the operation in the Modality-Aware module, $\tilde{S}_{6}$ is reshaped to $\mathbb{R}^{B \cdot N_6 \times 4  \times \frac{C_6}{4}}$ and Eq~\ref{eq:MHA} are applied. The output features are concatenated together along channels to generate ${F}'_{6}$.

\subsection{Decoder}
In the decoding stage, we first fold ${F}'_{j}$ back to a $4D$ feature map $\mathbb{R}^{d_j\times w_j \times h_j \times C_j}$. Subsequently,  with a $3 \mathrm{D}$ convolution and $2 \times$ upsampling operation, the resolution of the feature maps is increased by a factor of 2, and the outputs are concatenated with the outputs of the previous stage, a full resolution feature map is obtained and then converted to the final segmentation outputs by a sigmoid activation function. 

The soft Dice loss function~\citep{milletari2016v} is adopted as follows:
\begin{equation}
\mathcal{L}(G, P)=1-\frac{2}{J} \sum_{j=1}^J \frac{\sum_{i=1}^I G_{i, j} P_{i, j}}{\sum_{i=1}^I G_{i, j}^2+\sum_{i=1}^I P_{i, j}^2}
\end{equation}
where $I$ and $J$ denote the number of voxels and classes, respectively. For class $j$ at voxel $i$, $P_{i, j}$  denote the prediction of our model, and  $G_{i, j}$ is the ground truth.

\begin{table*}[htp]
    \centering
    \caption{Quantitative comparison on BraTS 2021 dataset with respect to Dice score and 95\% Hausdorff Distance. ET,
WT and TC denote Enhancing Tumor, Whole Tumor and Tumor Core respectively.}
    \label{tab:dataset_result} 
    \renewcommand\arraystretch{1.1}
    \resizebox{\textwidth}{!}{
    \begin{tabular}{c |c c| c c c c c c c c c}
    \Xhline{1pt}
    \multirow{2}{*}{Methods} & \multirow{2}{*}{\makecell{Param\\(M)}} & \multirow{2}{*}{\makecell{FLOPs\\(G)}} & \multicolumn{4}{c}{Dice$\uparrow$} & & \multicolumn{4}{c}{HD95$\downarrow$} \\
         \cline{4-7} \cline{9-12}  
     & & &ET & WT &TC &Avg & &ET & WT &TC &Avg\\
    \Xhline{1pt} 
   UNETR~\citep{hatamizadeh2022unetr} &71.31 &1159.0 &0.852 &0.922 &0.866 &0.880 & &12.26 &7.78 &7.73 &9.26 \\
   SegTransVAE~\citep{pham2022segtransvae} &44.72 &400.7 &0.862 &0.925 &0.899 &0.895 & &10.59 &7.71 & 5.88 &8.06  \\
    SwinUNETR~\citep{hatamizadeh2022swin}  &62.19&774.8 &0.871 &0.925 &0.899 &0.897 & &11.06 &7.62 &6.86 &8.51 \\
    \hline
    MMEF-nnUNet~\citep{huang2022evidence}  &76.85 &208.1 &0.872 &0.928 &0.900 &0.900 & &9.68 &8.29 &\textbf{5.10} &8.29 \\
    CKD-TransBTS~\citep{lin2022ckd}&- &-  & 0.885 &0.933 &0.901 &0.906 & &5.93 &6.20  &6.54 &6.22  \\
    NestedFormer~\citep{xing2022nestedformer}  &\bf{10.57} &206.9 &0.882 &0.932 &0.909 &0.908 & &7.14 &7.88 &5.43 &6.81 \\
    \Xhline{1pt}
    MASM (Ours) &24.89 &\bf{160.1}
 &\textbf{0.888} &\textbf{0.934} &\textbf{0.912} &\textbf{0.912} & &\textbf{5.72} &\textbf{5.94} &5.40 &\textbf{5.65}\\
    \Xhline{1pt}
    \end{tabular}
    }
\end{table*}
\begin{table*}[!h]
	\centering
 	\caption[t]{Mean Dice score for  Enhancing Tumor, Whole Tumor, and Tumor Core in terms of five-fold cross-validation benchmarks. ${\dag}$ denotes our implementation and * means we citep results from the original paper}
        \renewcommand\arraystretch{1}
	\label{tab:five_fold}
	\begin{tabular}{l|c|c|c|c|c|c|c|c|c|c|c|c}
		\hline 
        &\multicolumn{3}{c|}{MASM} &
	\multicolumn{3}{c|}{NestedFormer$^{\dag}$}
	& \multicolumn{3}{c|}{SwinUNETR$^*$} &
	\multicolumn{3}{c}{nnU-Net$^*$}\\ \hline
		Dice & ET & WT & TC & ET & WT & TC & ET & WT & TC & ET & WT & TC  \\ \hline
		Fold 0 &\textbf{0.891} &\textbf{0.933} &\textbf{0.915} &0.867 & 0.927 & 0.904 & 0.876 & 0.929 & 0.914 & 0.866 & 0.921 & 0.902 \\
		Fold 1 &0.901 &0.936 &\bf{0.919} &0.899 & 0.934 &0.915 & \bf{0.908} & \bf{0.938} & \bf{0.919} & 0.899 & 0.933 & \bf{0.919} \\
		Fold 2 &\bf{0.891} &\bf{0.933} &0.918 & 0.890 & 0.931 & 0.915 & \bf{0.891} & {0.931} & \bf{0.919} & 0.886 & 0.929 & 0.914 \\
		Fold 3 &\bf{0.890} &0.933 &0.918 &0.889  &0.930  &0.916  & \bf{0.890} & \bf{0.937} & \bf{0.920} & 0.886 & 0.927 & 0.914  \\
		Fold 4 &\bf{0.891} & \bf{0.935} &\bf{0.919} &0.885   &0.933  &0.917 & \bf{0.891} & 0.934 &0.917 & 0.880 & 0.929 & {0.917} \\ \hline
            Avg. & \bf{0.892} & \bf{0.934} & \bf{0.917} & 0.886 & 0.931 & 0.913 & 0.891 & 0.933 & \bf{0.917} &  0.883 & 0.927 & 0.913 \\
        \bottomrule
	\end{tabular}
\end{table*}

\section{Experiments}
\subsection{Dataset}
The BraTS 2021 dataset~\citep{menze2014multimodal,baid2021rsna,bakas2017advancing} is a public brain tumor segmentation dataset, including 1251 and 219 cases in the training and validation set, respectively. Each case contains four MRI modalities: a) T1-weighted, b) T1 contrasted-enhanced, c) T2-weighted, and d) T2 Fluid-attenuated Inversion Recovery (T2-FLAIR), which are rigidly aligned and resampled to the same resolution. The data were collected from multiple centers with different MRI scanners, and the labels in the training set were annotated by experts~\citep{bakas2017advancing,bakas2018identifying}.  The task of the dataset is to segment regions of brain tumors (i.e., whole tumor (WT), tumor core (TC), and enhancing tumor (ET)). Since the segmentation labels of the validation set are not publicly available, we adopt the training set for all the experiments.

\subsection{Implementation Details}
Our framework is implemented using PyTorch on an NVIDIA GTX 3090 GPU. We adopt U-Net with six layers as the backbone of our architecture. The channels $C_j$ of each layer are $\{96,128,192,256,384,512\}$. 
The initial learning rate is $1\times10^{-4}$, and we decay it following the learning rate scheduling strategy of \citep{hatamizadeh2022swin}. To ensure consistency with the experiment settings of previous work~\citep{xing2022nestedformer,hatamizadeh2022swin,zhang2021modality}, each volume is cropped into patches with a size of $128 \times 128 \times 128$  and normalized to have zero mean and unit standard deviation according to non-zero voxels. Random mirroring, shift, and scale are applied for data augmentation. To gauge the performance, we employ the Dice score and 95\% Hausdorff Distance (HD95) as evaluation metrics.

\subsection{Comparison with SOTAs}
We conduct experiments on a widely used split~\citep{peiris2022robust} where the 1251 MRI scans are split into 834, 208, and 209 for training, validation, and testing, respectively. To demonstrate the effectiveness, we compare the performances of our model with a wide range of state-the-art models, including universal models (UNETR~\citep{hatamizadeh2022unetr}, SegTransVAE~\citep{pham2022segtransvae} and SwinUNETR~\citep{hatamizadeh2022swin}) and models designed for multi-modal imaging (MMEF-nnUNet~\citep{huang2022evidence}, CKD-TransBTS~\citep{lin2022ckd} and NestedFormer~\citep{xing2022nestedformer}). For a fair comparison, we adopt the results from the original papers and the result of NestedFormer is implemented by ourselves.

As illustrated in Table~\ref{tab:dataset_result}, MASM, with moderate model size and slight computation, can outperform all the state-of-the-art methods across all metrics and achieve the best segmentation performance. It can be observed that models with specific designs for multi-modal imaging attain a notable improvement compared to the universal models. The results indicate that exploring multi-modal features and dependencies is conducive to the tumor segmentation of MRI scans. Instead of considering single-modality spatial coherence and cross-modality coherence at high levels (i.e., NestedFormer), MASM introduces a more reasonable architecture for multi-scale features that is conformable to the property of each modality. Although with more parameters, it shows MASM not only improves the Dice score and HD95 score but also lowers the computations significantly with around 1/4 of the FLOPs. Such improvements demonstrate that our model can effectively learn multi-modal features and accurately identify the relationship between modalities. 



\begin{figure}[htb]
\centering
\scalebox{1}{
\includegraphics[width=\linewidth]{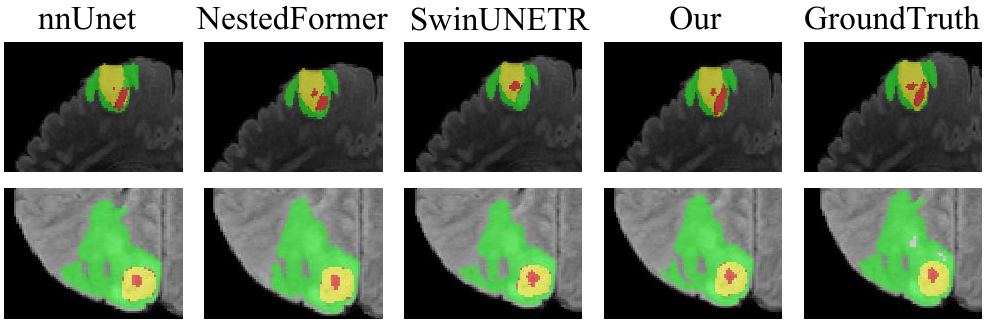}
}
\caption{Visualizations of comparisons with other methods.}
\label{fig:r2}
\end{figure}

To further evaluate our method, we compare our method in the cross-validation split following~\citep{hatamizadeh2022swin} with several methods (i.e., NestedFormer, SwinUNETR and nnUNet). The quantitative results are presented in Table~\ref{tab:five_fold}, where our model outperforms the previous methods across all five folds. We conducted a significant test with SwinUNETR, which shows remarkable results among previous work. The P-values for ET, WT, and TC are $\textbf{0.0001}$, $\textbf{0.0005}$ and $\textbf{0.02}$, respectively. The results demonstrate the effectiveness of our model even with only a modest improvement improvement in Dice scores. Moreover, 
We include additional visual comparison results in Figure~\ref{fig:r2}.

\subsection{Ablation Study} 
To fully analyze our proposed modules, we conduct ablation studies on the five-fold cross-validation. 

\begin{table}[htp]
\centering
\caption{Ablation study for proposed modules.}
\vspace{-5pt}
\label{tab:ablation_result}
\renewcommand\arraystretch{1.1}
\resizebox{\linewidth}{!}{
\begin{tabular}{c c |c c c | c c}
    \Xhline{1pt}
     \multicolumn{2}{c|}{Module} 
     &\multicolumn{3}{c|}{Dice$\uparrow$} & \multicolumn{2}{c}{Complexity$\downarrow$} \\
    \cmidrule{1-2} \cmidrule(lr){3-5} \cmidrule{6-7}  
     Aware &Shift &ET & WT &TC  &Param(M) &FLOPs(G)\\
    \Xhline{1pt}
     &  &0.868  &0.922  &0.903 &16.18 &148.75\\
    \hline
     {\checkmark} &  &0.879  &0.929  &0.911  &24.36 &160.12\\
    \hline
      &{\checkmark}  &0.876  &0.927  &0.908  &16.72 &149.14\\
    \hline
     {\checkmark} &{\checkmark}  &\textbf{0.892}  &\textbf{0.934}  &\textbf{0.917}  &24.89 &160.14\\
    \Xhline{1pt}
    \end{tabular}
    }
\vspace{-5pt}
\end{table}
\subsubsection{Effect of proposed modules} 
\begin{figure}[htb]
\centering
\scalebox{1}{
\includegraphics[width=\linewidth]{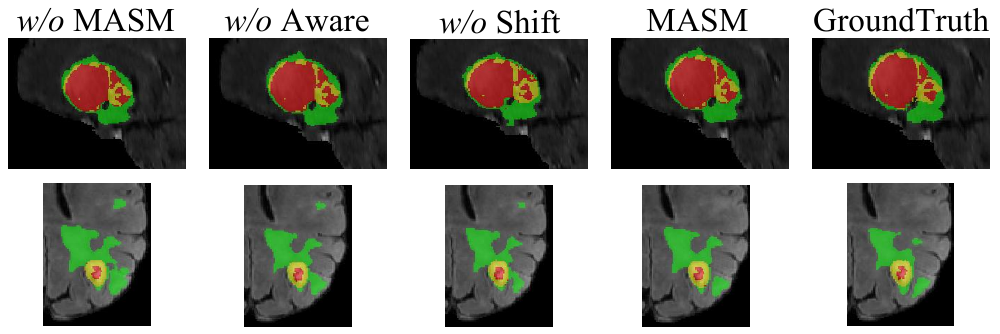}
}
\vspace{-10pt}
\caption{Visualizations of the results of designed modules.}
\vspace{-5pt}
\label{fig:r1}
\end{figure}

To evaluate the effectiveness of our method, we conduct ablation studies for critical components (i.e., Modality-Aware and Modality-Shift). Table~\ref{tab:ablation_result} summarizes the average results on the five-fold cross-validation benchmarks for the variants. We first remove the  Modality-Aware and Modality-Shift modules in our MASM as the baseline in our experiments. Then, we apply Modality-Aware in all layers and Modality-Shift in the last three layers, respectively. One thing worth noticing is that the baseline differs from nnU-Net~\citep{isensee2021nnu}, where the multi-modal images are separated and fed to one single encoder sharing parameters. As can be seen, Modality-Shift boosts performance with a margin (e.g., 0.868 $\rightarrow$ 0.876) in Dice score for Enhanced tumor, and Modality-Aware brings a more considerable improvement (e.g., 0.868 $\rightarrow$ 0.879). The performance gain of the Modality-Aware module and the Modality-Shift module demonstrates that using our proposed modules helps construct the relationship information and enhance the dependency information among modalities. The reason why the combination of the Modality-Aware and Modality-Shift modules leads to significant improvement is that the design of Modality-Aware is aligned with the experience of radiologists in routine diagnosis and the highly non-linear relationships of the high-level features can be modeled by both spatial and channel-wise attention in the Modality-Shift module. Moreover, it is observed that the shift operation does not cause much increment of model parameters and computation. To validate the effectiveness of the two modules intuitively, we visualize results in Figure~\ref{fig:r1}.
\vspace{-5pt}

\subsubsection{Effect of different backbones} 
In previous work, different backbones were typically applied to medical image segmentation tasks, such as UNETR and SwinUNETR.  To further evaluate the impact of our proposed modules, we apply Modality-Aware and Modality-Shift on different backbones. We compare variations of different backbones as shown in Table~\ref{tab:ablation_result1}. The modules are integrated into the specific layer of the framework where skip-connection is employed. It can be seen that replacing backbones does not boost performance. This could be partially attributed to the complex relationship modeling capability of the transformer-based backbone. It might degrade the interaction among multi-modal features. Moreover, comparing with the first three rows, we confirm that the proposed module is beneficial to medical multi-modal image segmentation.

\begin{table}[htp]
\centering
\vspace{-5pt}
\caption{Ablation study for different backbones. The results are implemented by ourselves.}
\vspace{-5pt}
\label{tab:ablation_result1}
\renewcommand\arraystretch{1.1}
\begin{tabular}{c |c c c c}
    \Xhline{1pt}
     \multirow{2}{*}{Backbones}  &\multicolumn{4}{c}{Dice$\uparrow$} \\ \cline{2-5}
       &ET & WT &TC & Avg\\
    \Xhline{1pt}
    UNETR &0.867	&0.921	&0.892 &0.893\\
    \hline
    SwinUNETR &0.861 &0.927 &0.898 &0.895 \\
    \hline
    UNet & 0.866 &0.924 &0.906 &0.898\\
    \hline
     MASM-UNETR  &0.881  &0.926  &0.905 &0.904\\
    \hline
     MASM-SwinUNETR  &0.883  &0.929  &0.908 &0.906  \\
    \hline
     MASM-UNet  &\textbf{0.892} &\textbf{0.934} &\textbf{0.917} &\textbf{0.914}  \\
    \Xhline{1pt}
    \end{tabular}
\vspace{-10pt}
\end{table}

\subsubsection{Effect of different ratios of two modules} 
In the encoder part, the input image first passes through a series of convolutional layers to capture fine-grained details and edge information, which can be regarded as low-level features. As the layer goes deeper, the receptive field becomes larger. These features represent higher-level abstractions. We design the Modality-Aware and the Modality-Shift modules for low-level and high-level features, respectively. To investigate the impact of the different ratios of two modules, we apply the different ratios of two modules to $6$ block features, the results are shown in Table~\ref{tab:layer}.

\begin{table}[htp]
\centering
\vspace{-5pt}
\caption{Analysis for different ratios of two modules.}
\vspace{-5pt}
\label{tab:layer}
\renewcommand\arraystretch{1.1}
\begin{tabular}{c |c c c c}
    \Xhline{1pt}
     \multirow{2}{*}{MA:MS}  &\multicolumn{4}{c}{Dice$\uparrow$} \\ \cline{2-5}
       &ET & WT &TC & Avg\\
    \Xhline{1pt}
     3:3  &0.870  &0.929  &0.905 &0.901\\
    \hline
     4:2  &0.889  &0.932  &0.911 &0.910  \\
    \hline
     5:1  &\textbf{0.892} &\textbf{0.934} &\textbf{0.917} &\textbf{0.914}  \\
    \hline
     6:0  &0.888 &0.930 &0.911 &0.910 \\
    \Xhline{1pt}
    \end{tabular}
\vspace{-10pt}
\end{table}

\subsection{Qualitative Analysis}
\begin{figure}[htp]
\centering
\scalebox{1}{
\includegraphics[width=\linewidth]{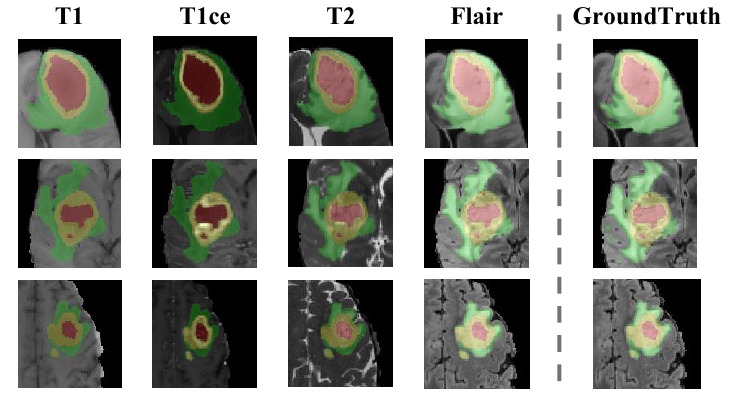}
}
\caption{The visual comparison results on BraTs 2021. Segmentation examples of the predicted labels (\textcolor{red}{ET},  \textcolor{green}{WT}, \textcolor{yellow}{TC}) are overlaid on T1, T1ce, T2, and FLAIR MRI axial slices in each row. The right column is the Ground Truth.}
\label{fig:visual}
\vspace{-10pt}
\end{figure}

 To better understand the effectiveness of our model, we also visualize several segmentation results in Figure~\ref{fig:visual}. Intuitively, the
results predicted by MASM are accurate and robust, which shows better alignment with ground truth. As the figure shows, owing to the employment of the proposed Modality-Shift and Modality-Aware, our model is able to effectively fuse multi-modal MRIs and accurately segment brain tumors and peritumoral edema, even for small regions.

\section{Conclusion}
In this work, we present a simple yet effective approach MASM for segmenting multi-modal volumetric medical images. The key insight is to construct and identify the relationship across modalities accurately. Our proposed model uses a CNN-based U-shaped network as the encoder and decoder. Furthermore, MASM incorporates the Modality-Aware and Modality-Shift modules for learning intra- and inter-modality dependencies and is able to capture representations at multiple scales efficiently and effectively. Experimental results on the BraTS 2021 dataset validate the effectiveness of our approach. Ablation studies also prove the potential of the proposed parts. Overall, we hope this architecture can shed novel insights into learning from multi-modal medical images. More applications of MASM in medical image segmentation will be considered in future work.

\bibliographystyle{named}
\bibliography{ijcai24}

\end{document}